\documentclass[a4paper,conference]{IEEEtran}
\usepackage{cite}
\ifCLASSINFOpdf
\else
\fi
\usepackage{amsmath}
\usepackage{algorithmic}
\usepackage{array}

\usepackage{graphicx}

\begin{document}
\title{Global Image Sentiment Transfer}

\author{
\IEEEauthorblockN{Jie An}
\IEEEauthorblockA{Department of Computer Science\\
University of Rochester\\
Rochester, NY, USA\\
Email: jan6@cs.rochester.edu}
\and
\IEEEauthorblockN{Tianlang Chen}
\IEEEauthorblockA{Department of Computer Science\\
University of Rochester\\
Rochester, NY, USA\\
Email: tchen45@cs.rochester.edu}
\and
\IEEEauthorblockN{Songyang Zhang}
\IEEEauthorblockA{Department of Computer Science\\
University of Rochester\\
Rochester, NY, USA\\
Email: szhang83@ur.rochester.edu}
\and
\IEEEauthorblockN{Jiebo Luo}
\IEEEauthorblockA{Department of Computer Science\\
University of Rochester\\
Rochester, NY, USA\\
Email: jluo@cs.rochester.edu}
}


\author{\IEEEauthorblockN{Jie An\IEEEauthorrefmark{1},
Tianlang Chen\IEEEauthorrefmark{1},
Songyang Zhang\IEEEauthorrefmark{1} and
Jiebo Luo\IEEEauthorrefmark{1}}
\IEEEauthorblockA{\IEEEauthorrefmark{1}Department of Computer Science\\
University of Rochester\\
Rochester, NY, USA\\ Email: \{jan6,tchen45,jluo\}@cs.rochester.edu, szhang83@ur.rochester.edu}}


\maketitle

\begin{abstract}
Transferring the sentiment of an image is an unexplored research topic in the area of computer vision. This work proposes a novel framework consisting of a \emph{reference image retrieval} step and a \emph{global sentiment transfer} step to transfer sentiments of images according to a given sentiment tag. The proposed image retrieval algorithm is based on the SSIM index. The retrieved reference images by the proposed algorithm is more content-related against the algorithm based on the perceptual loss. Therefore can lead to a better image sentiment transfer result. In addition, we propose a global sentiment transfer step, which employs an optimization algorithm to iteratively transfer sentiment of images based on feature maps produced by the Densenet121 architecture. The proposed sentiment transfer algorithm can transfer the sentiment of images while ensuring the content structure of the input image intact. The qualitative and quantitative experiment demonstrates that the proposed sentiment transfer framework outperforms existing artistic and photorealistic style transfer algorithms in making reliable sentiment transfer results with rich, fine, and exact details.
\end{abstract}


\IEEEpeerreviewmaketitle

\section{Introduction} \label{sec:intro}
Transferring the sentiment of an image is still an unexplored research topic. Comparing with the existing well-known tasks such as two-domain image-to-image translation \cite{huang2018multimodal,zhu2017unpaired,lee2018diverse,tang2019cycle} (\emph{e.g.} winter $\rightarrow$ summer, cat $\rightarrow$ dog) and image style transfer (e.g. artistic style transfer, photorealistic style transfer), image sentiment transfer focuses on modifying the image from a higher-level aspect to change it overall feeling to people. For example, without modifying the content, a family portrait can be transferred to be a more positive picture. The transferred one may give people a feeling of warmth and thus be more valuable to be kept. As we live in an age of pressures, we argue that this research topic is significant with its strong potential to decorate people's life.

\begin{figure}[t]
    \centering
    \includegraphics[width=\linewidth]{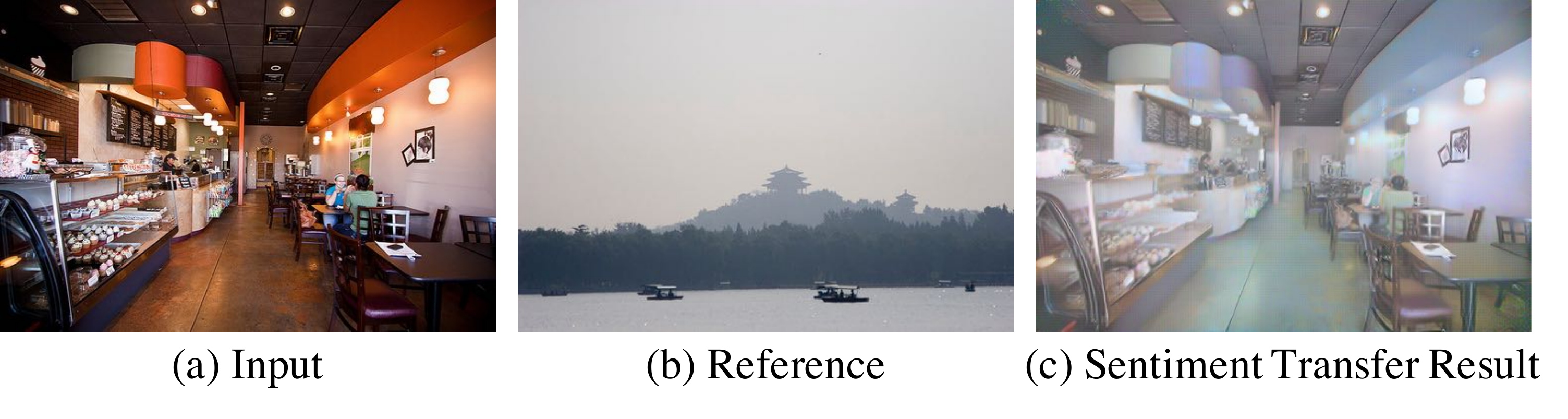}
    \label{fig:intro}
    \caption{\textbf{Failed sentiment transfer case with a content-unrelated image as the reference image.} The generated image is not photorealistic.}
\end{figure}

Intuitively, image sentiment is an abstract concept. Compared with the two-domain image-to-image translation that commonly has a definite pattern to accomplish the transfer between two domains (\emph{e.g.} cat $\rightarrow$ dog, horse $\rightarrow$ zebra), there are enormous ways to transfer an image to a specific sentiment. To make the image transfer controllable, a reference image should be fed into the model as guidance. Considering its similarity to the image style transfer task, we can leverage existing image style transfer models to perform reference-guided image sentiment transfer. However, it is nontrivial to implement this design because of the poorer compatibility between the input image and the reference one for image sentiment transfer.
Moreover, directly using existing artistic and photorealistic style transfer models generally fails to create visually pleasing results in terms of detail preservation and artifacts/distortions elimination. Compared with the image style transfer that an artistic/photorealistic style can be indiscriminately added to any input images, the sentiment transfer between two content-unrelated images is risky. Given the example of Fig.~\ref{fig:intro}, the sentiment transfer result lakes photorealism due to the reference image does not bring any content-related reference information to the input image. 

Considering this, we propose a high-performance image sentiment transfer framework that starts with image retrieval. Given an input image and a sentiment tag provided by the user, instead of randomly sampling a reference image that contains the input sentiment tag, we retrieve the most suitable reference image based on the structural information of the input image. Leveraging structural similarity (SSIM), the framework significantly constructs the content relation between the input and the reference image. In Section~\ref{sec:exp}, we demonstrate that this image retrieval step is crucial to improve the performance of image sentiment transfer. 

To transfer the sentiment of the retrieved reference image to the input image, we design a novel global image sentiment transfer algorithm. Inspired by the image style transfer algorithm by Gatys $et al.$~\cite{gatys2015neural}, we use an optimization algorithm on deep features by the neural network pre-trained on the ImageNet~\cite{krizhevsky2012imagenet} dataset to iteratively transfer the sentiment of the reference image to the input image. Different from existing style transfer algorithms, our method adopts the Densenet121 architecture as the feature extractor instead of the widely-used VGG19 architecture. We emperically find that Densenet121 architecture outperforms VGG19 architecture in terms of fine detail preservation and artifacts/distortions elimination. Therefore is more suitable to make sentiment transfer where the produced image should be photorealistic.

Our main contributions are summarized as follows:
\begin{itemize}
    \item We are the first to explore the task of image sentiment transfer. We present an effective two-step framework for the task by image retrieval and reference-guided image sentiment transfer.  
    \item We introduce an effective reference image retrieval algorithm based on SSIM index, which can achieve better results compared with other methods in finding content-related reference images.
    \item We propose a global sentiment transfer algorithm based on the Densenet121, which can transfer the sentiment/style of an image while preserving fine details of the image.
\end{itemize}

\section{Related Work}

% \subsection{Sentiment Applications}

Visual sentiment understanding has been explored for many years. 
Most existing works focus on the visual sentiment classification tasks.
To perform accurate classification for images with different sentiments, low-level features, like color~\cite{alameda2016recognizing,sartori2015s,machajdik2010affective}, texture~\cite{machajdik2010affective}, and shape~\cite{lu2012shape} has been studied in early years.
Later on, mid-level composition~\cite{machajdik2010affective}, sentributes~\cite{yuan2013sentribute}, principles-of-art features~\cite{zhao2014exploring}, high-level noun-adjective pairs (ANP)~\cite{borth2013large} are also been considered.
Most recently, due to the rapid development of the convolution neural network (CNN) for extracting visual features, many approaches turn to work on the CNN-based sentiment recognition. 
Some of them working on the noisy data during the training process~\cite{yang2018retrieving,yang2017joint,you2015robust}, while some of them exploring the visual sentiment in region level~\cite{yang2018weakly,song2018boosting,zhao2019pdanet,rao2019multi,you2017visual}.
However, compared with image sentiment classification, the other sentiment-related fields such as image sentiment generation/translation has not been well studied yet.

% \subsection{Style Transfer}

The most related tasks to ours are the image-to-image translation and the image style transfer.
Image-to-image translation targets at learning an image-to-image mapping from two different domains.
Early approaches need paired data to train the model and are essentially restricted to learn the deterministic one-to-one mapping~\cite{karacan2016learning,sangkloy2017scribbler,isola2017image}. This disables the generation of diverse outputted images.
CycleGAN~\cite{zhu2017unpaired} first proposes a cycle consistency loss to enable the model to be trained from the unpaired data. Following approaches like MUNIT~\cite{huang2018multimodal} and DRIT~\cite{lee2018diverse} further propose disentangled representations that enable the outputted images to be diverse.
On the other hand, our task is related to image style transfer. A great number of approaches are proposed for artistic style transfer~\cite{liao2017visual,huang2017arbitrary,gatys2016image,gatys2017controlling,kotovenko2019content} and photorealistic style transfer~\cite{li2018closed,luan2017deep,yoo2019photorealistic,bae2006two,an2019ultrafast}. 
Different from the above approaches, we focus on image sentiment transfer that requires a strong content relation between the input and the reference image. Therefore, we search the reference image based on the sentiment tag provided by the users instead of directly asking users to provide it.
The proposed global sentiment transfer algorithm is based on the work by Gatys $et al.$~\cite{gatys2016image}. However, our algorithm uses the Densenet121~\cite{huang2017densely} network architecture instead of the VGG19~\cite{simonyan2014very} as the feature extractor since we empirically find that Densenet121 can achieve a more faithful input detail preservation compared with the VGG19.

Searching a reference image by a given sentiment tag is related to the image retrieval task. 
Image retrieval aims at finding an image that is close to the given image.
%Earlier work propose numerous simple, shallow functions to directly measure the distance between raw images, such as SSIM~\cite{wang2004image}, FSIM~\cite{zhang2011fsim}, MSSIM~\cite{wang2003multiscale}.
Most recent works measure the perceptual loss of images by comparing the image features extracted from pre-trained convolution neural networks~\cite{zhang2018perceptual}. Images with more similar style effects generally have a low perceptual loss. 
Different from these works based on the perceptual loss, we find that the SSIM index~\cite{wang2004image} is more suitable to retrieve reference images for sentiment transfer since it mainly captures the similarity of images in terms of the content notions rather than focusing on stylization effects.

\begin{figure*}[t]
    \centering
    \includegraphics[width=\textwidth]{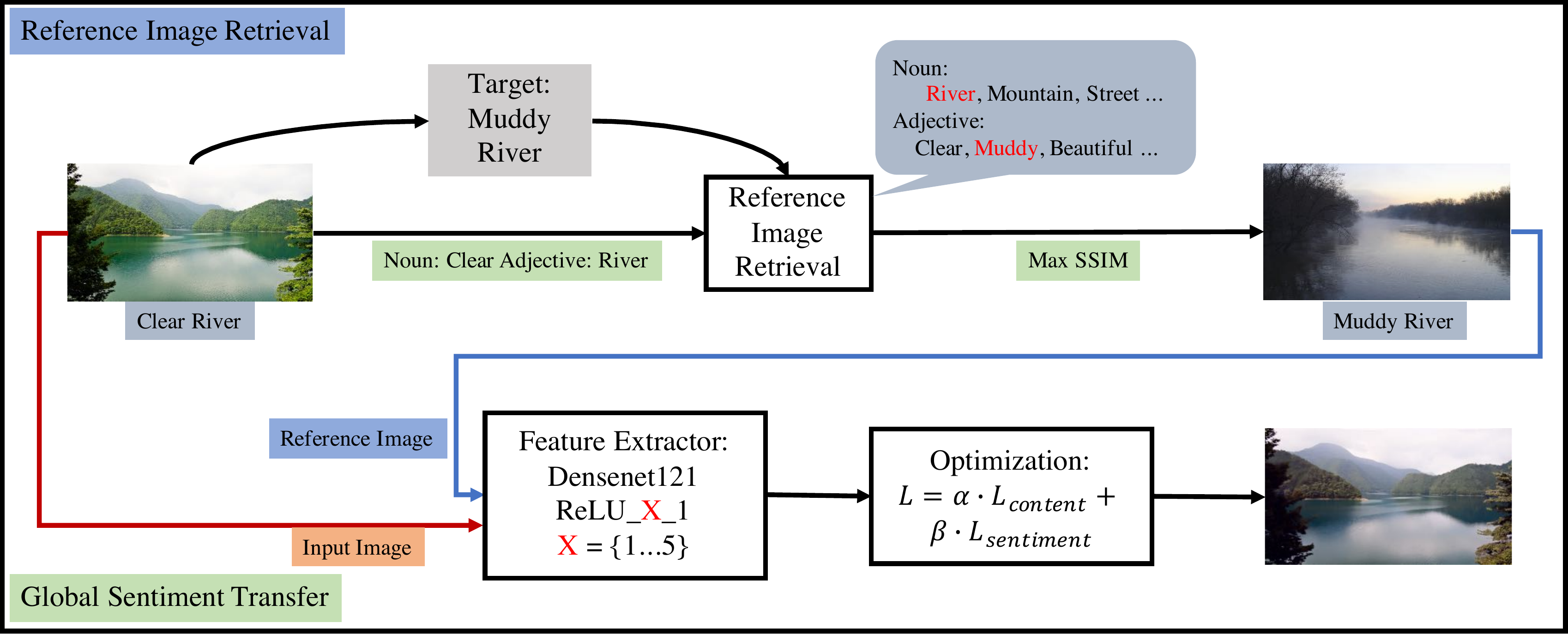}
    \label{fig:framework}
    \caption{\textbf{Framework of the proposed algorithm.} Our method consists of two parts: a reference image retrieve algorithm and a global sentiment transfer approach based on the retrieved reference image.}
\end{figure*}
\begin{figure*}[t]
  \includegraphics[width=\textwidth]{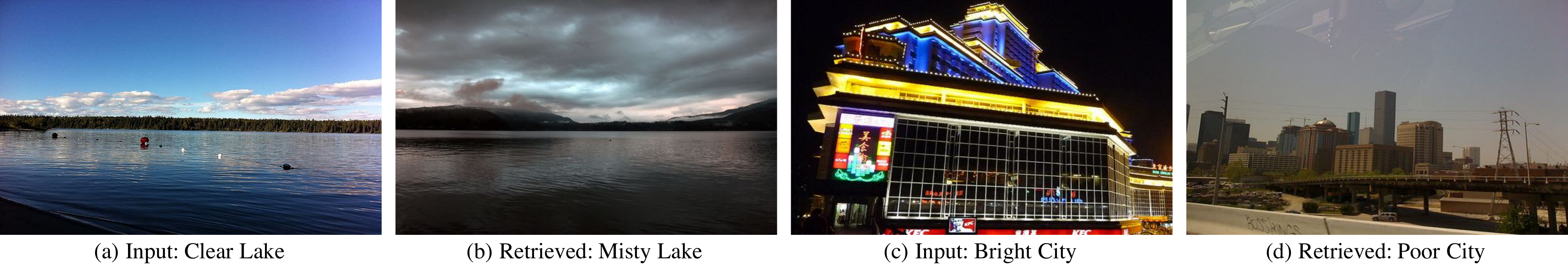}
  \caption{\textbf{Reference images retrieved by the proposed algorithm on VSO dataset.} The retrieved reference images have the same content with the input image but contain the opposite sentiments.}
  \label{fig:retrieve}
\end{figure*}
\section{Method}
To transfer the global sentiment of an image, we propose a method that consists of a \emph{reference image retrieve} and \emph{global sentiment transfer} step. Fig.~\ref{fig:framework} shows the framework of the proposed algorithm. Given an input image, we first retrieve a reference image according to the target sentiment tag. Then a global sentiment transfer algorithm is employed to transfer the sentiment of the input image to the reference. We describe the details of these two steps in the following part of this section.
\subsection{Reference Image Retrieve}
Retrieving a reference image is the initial step to make sentiment transfer. Given an input image, the proposed retrieval method aims at finding a reference image according to a sentiment tag given by the user. To facilitate the following global sentiment transfer step, the retrieved image should have a similar content compared with the input image but contains the feelings of the target sentiment.

To achieve this, we propose an image retrieval algorithm based on the Visual Semantic Odometry (VSO)~\cite{lianos2018vso} dataset. For each image in the VSO dataset, a noun-adjective pair is attached with it to describe the semantic content and its sentiment respectively. To retrieve a reference image according to a given sentiment tag, we first select a subset of the VSO dataset, where every image within the subset contains the content tag (noun) of the input image but the sentiment tag (adjective) of the given target, For example, in Fig.~\ref{fig:framework}, the input image has a noun-adjective pair of ``Clear River'' while images in the corresponding subset have a label of ``Muddy River''.

%Since the performance of the sentiment transfer algorithm significantly relies on the semantic similarity between the input and reference images, a more similar content structure can lead to a better sentiment transfer result. To explain this, we introduce an extreme example where the input image and the reference image are almost the same except having different colors. To make sentiment transfer on this input-reference pair, the sentiment transfer algorithm only needs to change the color of the input according to the target.

To find a reference image from the selected target subset which has the most similar semantic structure with the input image, inspired by~\cite{yoo2019photorealistic,an2019ultrafast}, we use the Structural Similarity Index (SSIM)~\cite{chen2011fast} between edge responses~\cite{xie2015holistically} of images to measure the semantic similarity between each image in the target subset and the input image. SSIM index is originally used by image/video quality assessment methods. We empirically find that SSIM is more suitable than the widely-used perceptual loss to measure the semantic similarity between two images. For every image in the target subset, we first compute the SSIM index between the evaluated and the input images and then pick-up the image with the highest SSIM index as the corresponding reference image to the input image.

%The proposed image retrieve algorithm is xxx. (a paragraph to show its benefits against existing methods.) 

\begin{figure*}[t]
  \includegraphics[width=\textwidth]{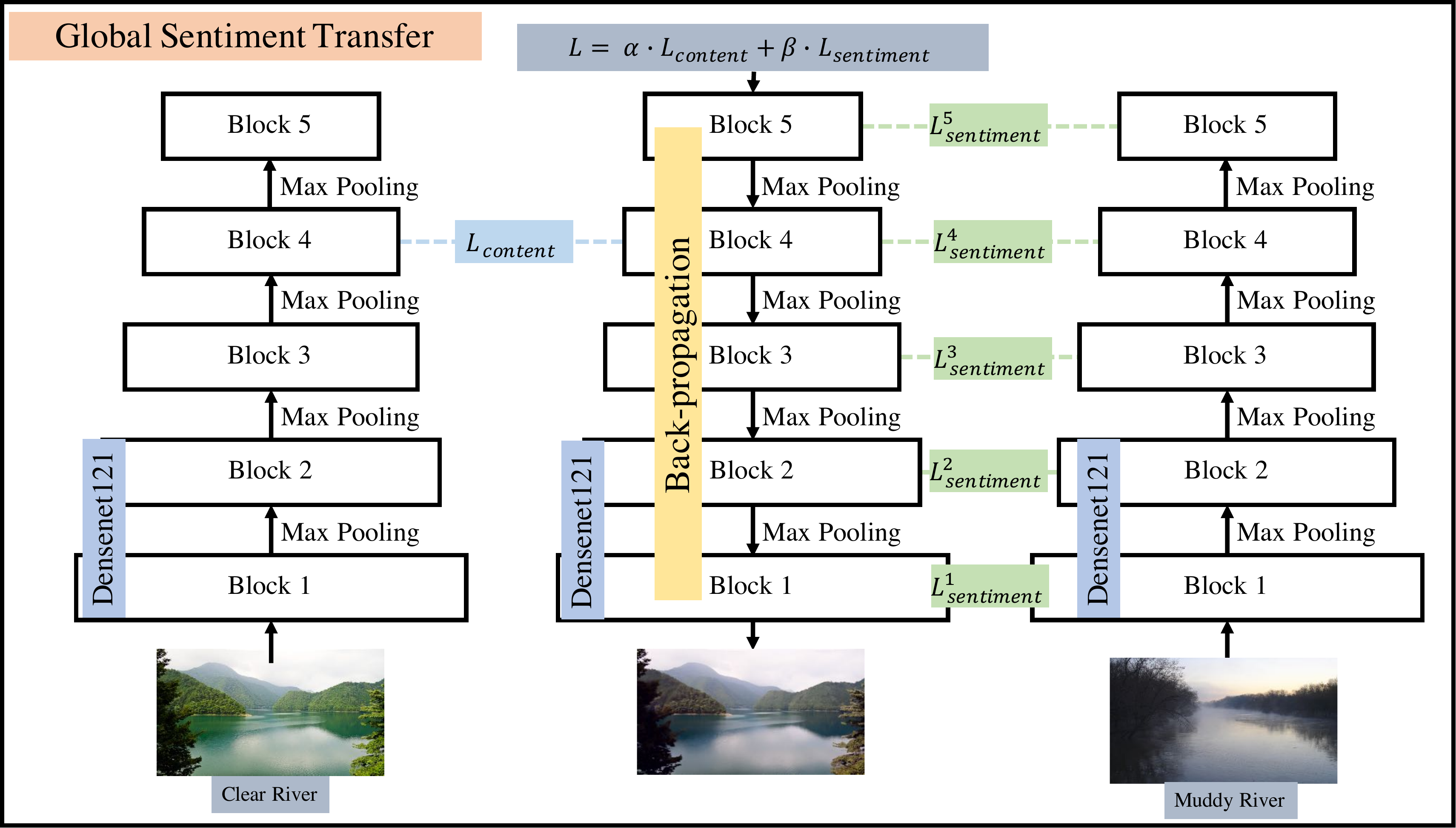}
  \caption{\textbf{Illustration of the proposed global sentiment transfer algorithm.} Here we use Densenet121 architecture as the backbone network.}
  \label{fig:transfer}
\end{figure*}

\subsection{Global Sentiment Transfer}
With the given input and a selected reference image that is structurally most similar to the input image, we propose a novel algorithm to transfer the sentiment of the input image according to the selected reference image. Our algorithm is based on an optimization method, which iteratively transfers the sentiment of images by minimizing two objectives on deep features. Here the first objective is to ensure the details of the input intact while the other one is to restrain the sentiment of the produced image similar with the reference image.

A high-quality sentiment transfer result should have a similar sentiment with the reference image while keeping the content details intact compared with the input image. The key challenge in sentiment transfer is to measure the sentiment similarity between two images. Inspired by Gatys et al.~\cite{gatys2015texture,gatys2015neural}, we adopt the Gram loss on deep features of the input and reference images produced by neural networks to measure the sentiment similarity. Such a Gram-based loss term is originally used to measure the style similarity. Since sentiment can be regarded as the abstract of the style, we borrow the Gram loss term to make sentiment transfer. Moreover, we compute the $l_2$ norm between features of the transferred and input images as the content-consistency loss.

\begin{figure}[t]
  \includegraphics[width=\linewidth]{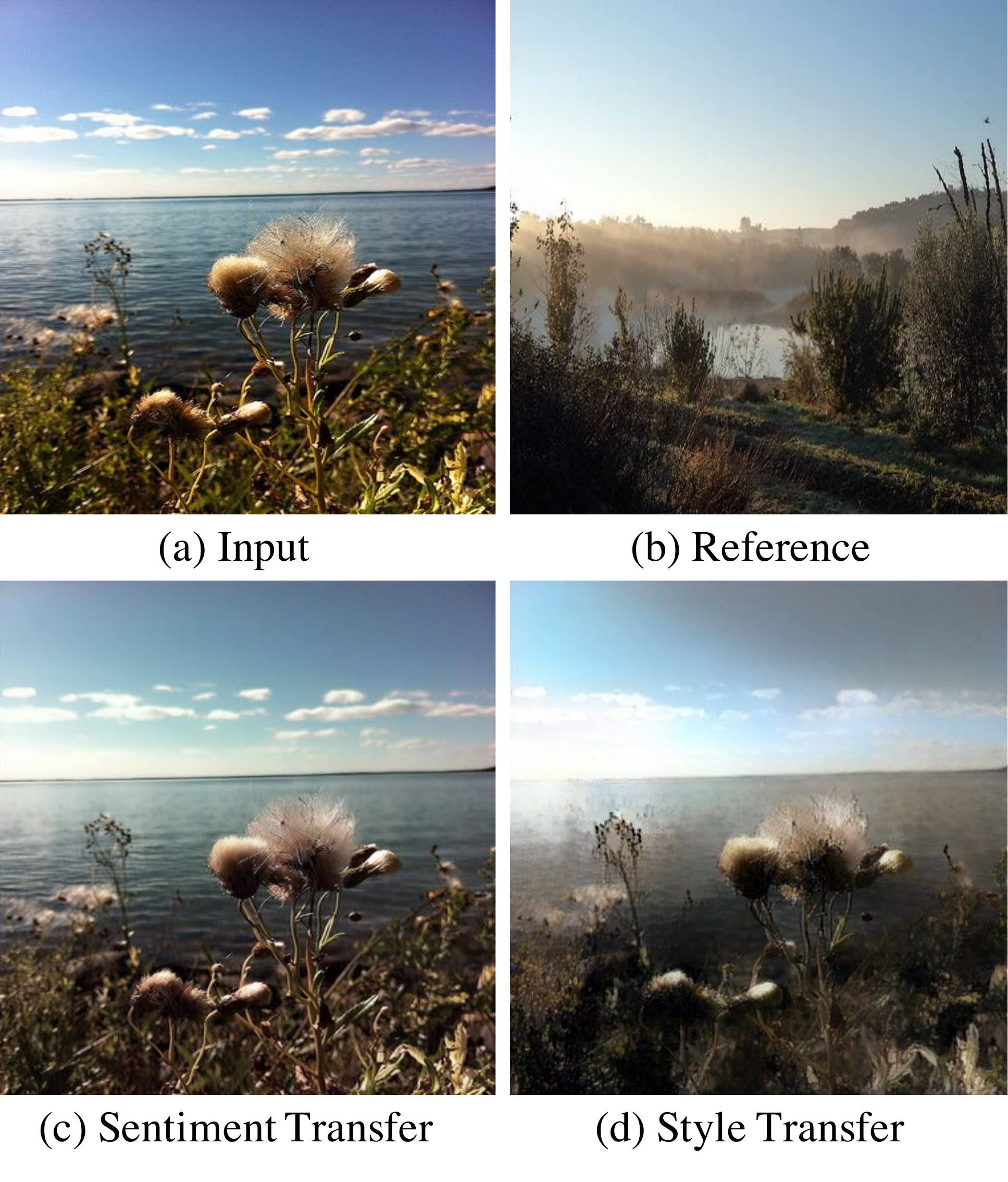}
  \caption{\textbf{Comparison between style transfer results and sentiment transfer results.} The style transfer results are produced by the VGG19 architecture while the sentiment transfer images are generated based on the Densenet121.}
  \label{fig:comp_style}
\end{figure}

The sentiment/style transfer results created by the above-mentioned loss terms heavily relies on deep features used to compute objectives. Those style transfer algorithms~\cite{gatys2015texture,Gatys2016,gatys2016preserving,risser2017stable,luan2017deep,li2017laplacian,li2017demystifying} all use features produced by the VGG19 network pre-trained on ImageNet dataset~\cite{krizhevsky2012imagenet}. However, the optimization algorithm based on the feature maps of VGG19 can inevitably change details of the content image. Please take the style transfer results of Gatys et al.~\cite{Gatys2016} shown in Fig.~\ref{fig:comp_style} for example. The style transfer algorithm based on features by VGG19 changes details of the sea, sky, and plants in the input image.

As illustrated in Fig.~\ref{fig:transfer}, the proposed algorithm adopts the Densenet121 network which is pre-trained on ImageNet dataset~\cite{krizhevsky2012imagenet} as the feature extractor. We empirically find that Densenet121 can achieve sentiment transfer while avoiding the damage to the detail of the input content. Based on the Densenet121, We first get deep features of the input and the reference image produced by the ReLU layers behind each pooling operator. Here we use $f_s^i,\ i\in\left\{1...5\right\}$ and $f_t^i,\ i\in\left\{1...5\right\}$ to denote feature maps of the input and reference image respectively, where $s$ denotes \emph{source} while $t$ represents \emph{target}. The use of Densenet121 has two main advantages: first, Densenet121 networks can ensure the content information intact while transferring sentiment from the reference to the input image. Second, Densenet121 has only half of the parameters of VGG19 (Densenet121: 6.952 $v.s.$ VGG19: 12.945). Therefore it is more time-efficient in creating sentiment transfer results.

The overall loss functions we used is,
\begin{equation}
  \mathcal{L} = \alpha \cdot \mathcal{L}_{content} + \beta \cdot \mathcal{L}_{sentiment},
  \label{eq:1}
\end{equation}
\begin{equation}
  \mathcal{L}_{content} = \| f^4 - f^4_s \|_2,
\end{equation}
\begin{equation}
  \mathcal{L}_{sentiment} = \frac{1}{5}\cdot\sum_{i=1}^5\| \mathrm{Gram}\left(f^i\right) - \mathrm{Gram}\left(f_t^i\right) \|_2,
\end{equation}
where $\mathrm{Gram}\left(f\right) = f^T\cdot f$, $f^i$ denotes the feature maps by the transferred images in Densenet121. All the feature maps $f$ has the shape of $C \times \left(H \times W\right)$, where $C$ denotes the channel number while $H,W$ represent height and width of $f$ respectively. Please note that here the transferred image is the variable in optimization process, $i.e.$, we iteratively alter an image to make its str look the same as the input image while having the sentiment of the reference image.

\section{Experiment}
\label{sec:exp}
In this section, we first discuss experimental settings. Then we compare the proposed image retrieval and global sentiment transfer algorithm against other image retrieval and image style transfer algorithms respectively. Finally, we demonstrate the effectiveness of the proposed algorithm by both visual and quantitative evaluation.
All the source code and the trained model will be made available to the public.

\begin{table}
    \caption{\textbf{Selected global sentiment datasets from VSO images}.}
    \centering
    \begin{tabular}{c|c}
    \hline
      Positive Sentiments& Negative Sentiments \\
      \hline
      Warm home& Dark room\\
      Clear river& Muddy water\\
      Clear water& Muddy river\\
      Clear mountain& Misty mountains\\
      Scenic mountain& Rough hill\\
      Clear lake& Misty lake\\
      Lovely city& Harsh landscape\\
      Bright city& Poor city\\
      Great city& \\
      \hline
\end{tabular}
\label{tab:sentiment}
\end{table}

\subsection{Experimental Settings}
\noindent\textbf{Global Sentiment Dataset.\ } To demonstrate the effectiveness of the proposed global sentiment transfer framework, we collect a few images from the VSO dataset, which have global sentiments. For example, an image with the description of ``beautiful bird'' should not be selected since ``bird'' is only a regional object. Table.~\ref{tab:sentiment} shows the noun-adjective pairs of the selected subsets we used in our experiment. We employ nine subsets with positive sentiments and eight subsets with negative sentiments.

\noindent\textbf{Global Sentiment Transfer Settings.\ } To transfer the sentiment of the selected reference image to the input image, we propose an optimization-based iterative method. As stated above, we use a Densenet121 pre-trained on the ImageNet dataset as the feature extractor. In optimization, we use Adam algorithm~\cite{kingma2014adam} to make optimization on the input image and the retrieved reference image. To balance the content and sentiment loss terms, we set $\alpha=1$ and $\beta=1,000,000$ in Equation~\ref{eq:1}. To get the sentiment transfer result for each input-reference pair, we run the optimization method for 500 iterations.

\subsection{Reference Image Retrieve}
\begin{figure}[t]
  \includegraphics[width=\linewidth]{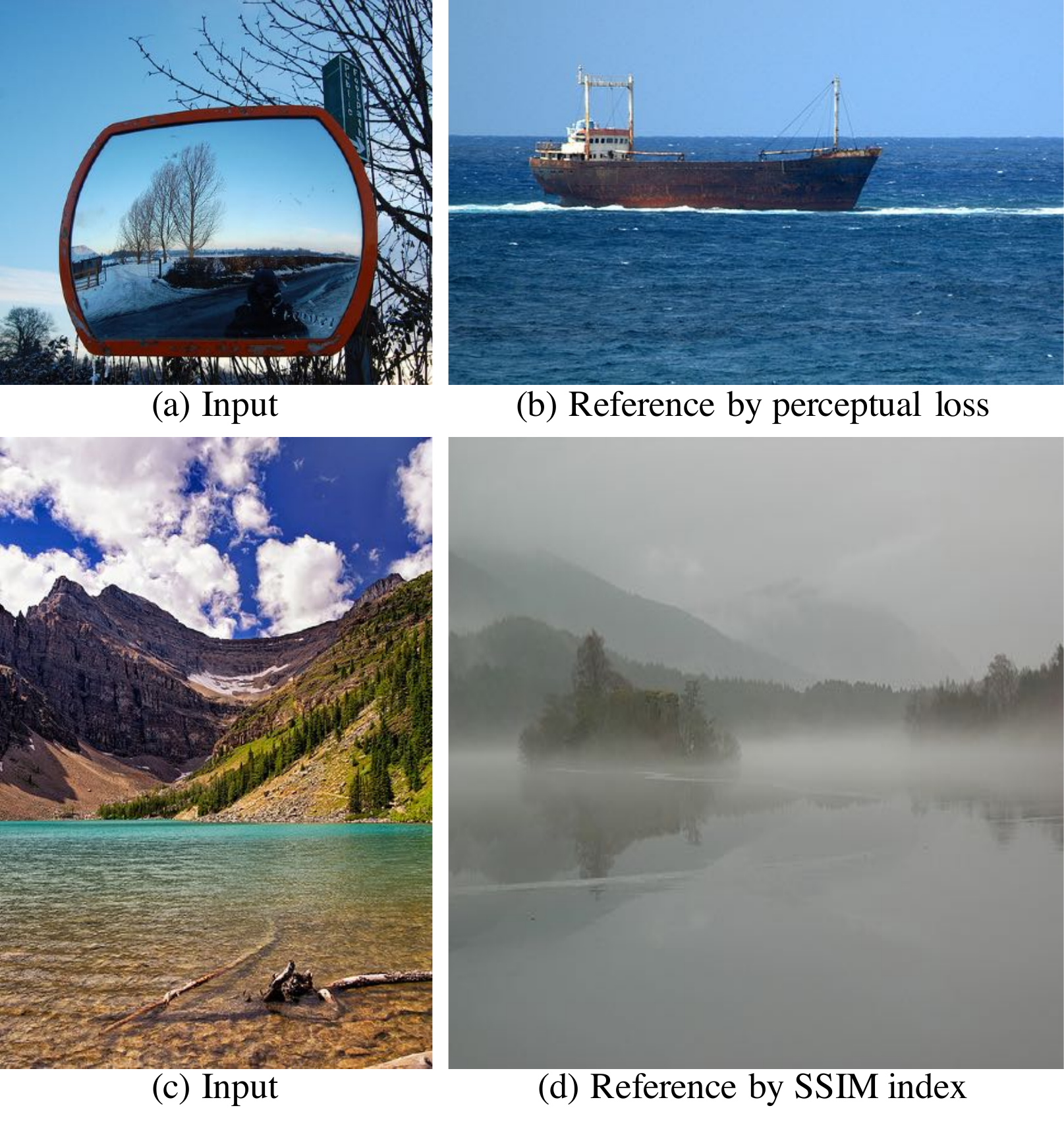}
  \caption{\textbf{Comparison between the proposed image retrieval method based on SSIM index and the image retrieval method based on the perceptual loss.} The proposed retrieval method focuses on finding images with most relevant content while the algorithm based on the perceptual loss computes the distance between images mainly in terms of the style similarity.}
  \label{fig:comp_retrieve}
\end{figure}

Fig.~\ref{fig:comp_retrieve} shows the image retrieval results based on the perceptual loss and SSIM index. Since the perceptual loss mainly measures the similarity between two images in terms of style effects, the retrieved image based on the perceptual loss generally has a similar sentiment/style but distinct content notations. On the contrary, the retrieved image produced by the algorithm based on the SSIM index contains the same content but a different sentiment compared with the input image. Please take Fig.~\ref{fig:comp_retrieve} for example. The retrieved image by the perceptual loss (b) has the same blue style compared with the input image (a). However, the content structure of (a) and (b) is completely different. On the contrary, the input (c) and the retrieved image by SSIM index (b) has a similar content structure. Generally, the global sentiment transfer algorithm would generate a better result if the reference image has a more similar content structure compared with the input image. Therefore, Fig.~\ref{fig:comp_retrieve} demonstrates that the proposed image retrieval method based on the SSIM index has a better performance compared with the algorithm based on the perceptual loss.

\subsection{Visual Comparison}
\begin{figure*}[t]
  \includegraphics[width=\linewidth]{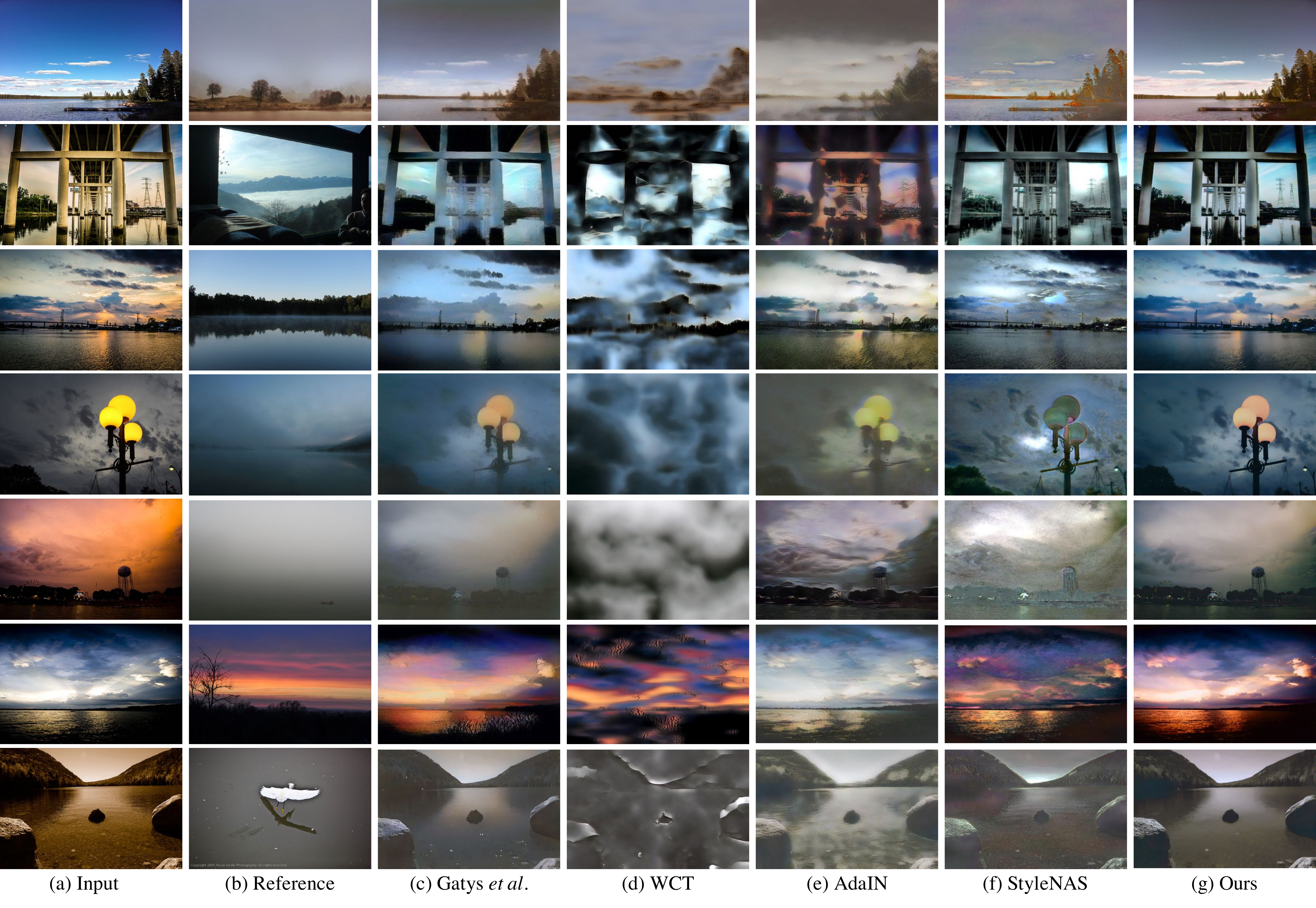}
  \caption{\textbf{Visual comparison between the results produced by the proposed global sentiment transfer algorithm and the state-of-the-art universal style transfer algorithms.} All the compared results are produced by running the officially-released code of the corresponding algorithm.}
  \label{fig:comp_visual}
\end{figure*}

\begin{table*}[t]
\caption{\textbf{Comparison of mean SSIM score on the validation set.} Higher SSIM score means better detail preservation ability.}
    \centering
    \begin{tabular}{lccccc}
        \hline
      Method\hspace{0.5cm} & \hspace{0.5cm}Gatys $et al.$~\cite{gatys2015neural}\hspace{0.5cm} & \hspace{0.5cm}WCT~\cite{li2017universal}\hspace{0.5cm} & \hspace{0.5cm}AdaIN~\cite{huang2017arbitrary}\hspace{0.5cm} & \hspace{0.5cm}StyleNAS~\cite{an2019ultrafast}\hspace{0.5cm} & Ours \\
      \hline
      SSIM$\uparrow$ & 0.7019 & 0.2443 & 0.5301 & 0.6653 & \textbf{0.8719} \\
      \hline
    \end{tabular}
    \label{tab:ssim}
\end{table*}

Since this work is the first global sentiment transfer algorithm for the arbitrary input image, we compare the result produced by our algorithm with state-of-the-art artistic~\cite{chen2016fast,huang2017arbitrary,li2017universal} and photorealistic~\cite{an2019ultrafast} style transfer algorithms to demonstrate the effectiveness of the proposed global sentiment transfer method. Other photorealistic style transfer algorithm such as~\cite{luan2017deep,li2018closed,yoo2019photorealistic} are not compared since these methods need a segmentation map or post process to assist the style transfer. Such pre or post processing steps are not needed by ours and the compared methods. To make a fair comparison, each compared style transfer algorithm adopts the retrieved image produced by our image retrieval algorithm based on the SSIM index as the reference image. Fig.~\ref{fig:comp_visual} shows the sentiment transfer results of our method and style transfer results by the state-of-the-art universal style transfer algorithms. The results by artistic style transfer algorithms ($e.g.$ StyleSwap\cite{chen2016fast}, WCT~\cite{huang2017arbitrary}, AdaIN~\cite{li2017universal}) usually has distorted content details, which is necessary to create artistic feelings but is not favored to make faithful sentiment transfer. The photorealistic style transfer algorithm~\cite{an2019ultrafast} can preserve the content information. However, it may create significant artifacts in images. Please take Fig.~\ref{fig:comp_visual} (f) for example. Fig.~\ref{fig:comp_visual} (g) shows the result produced by our global sentiment transfer algorithm. The results by our method successfully achieve the sentiment transfer while ensuring the content details intact. Moreover, the produced results have significantly fewer artifacts compared with the state-of-the-art photorealistic style transfer algorithms.

\subsection{Quantitative Comparison}
We quantitatively demonstrate the effectiveness of the proposed algorithm by computing the SSIM index between the input image and the produced result. Here the SSIM score measures the ability of the compared algorithm to preserve fine details of the content. We collect 46 input-reference image pairs to form a validation set. We get the sentiment/style transfer results with all the compared algorithms on this validation set. Table.~\ref{tab:ssim} shows the mean SSIM score of the compared algorithm on this validation set. The proposed global sentiment transfer algorithm has a higher mean SSIM score compared with other style transfer methods, which demonstrates that our method has a stronger ability to preserve fine details of the input.

\section{Conclusion}
In this work, we propose a high-performance global image sentiment transfer framework consisting of a reference image retrieval step and a global sentiment transfer step. In reference image retrieval step, we adopt the SSIM index instead of the widely-used perceptual loss to measure the structure distance of images, which can capture content similarity of images rather than style effects. In the global sentiment transfer step, we use the Densenet121 network pre-trained on the ImageNet dataset as the feature extractor and employ an image style transfer framework to iteratively transfer sentiment based on features produced by the Densenet121 architectures. Our qualitatively and quantitatively experiment demonstrates that the proposed algorithm outperforms existing style transfer algorithms in terms of sentiment transfer effects and input detail preservation.


%\section*{Acknowledgment}




%\bibliographystyle{IEEEtran}
%\begin{thebibliography}{1}
%\end{thebibliography}

\small
\bibliography{ms}
\bibliographystyle{IEEEtran}

\end{document}